%% file: emnlp2018.tex
\documentclass[11pt,a4paper]{article}
\usepackage[hyperref]{emnlp2018}
\usepackage{times}
\usepackage{latexsym}

\usepackage{helvet}  
\usepackage{courier}  
\usepackage{url}  
\usepackage{graphicx}  
\usepackage{tabularx}

\usepackage{todonotes}
\usepackage{comment}
\usepackage{caption}
\usepackage{array}
\usepackage{textcomp}
\usepackage{multicol}
\usepackage{mathtools}
\usepackage{amsmath}
\usepackage{amssymb}
\usepackage{multirow}
\usepackage{stfloats}
\usepackage{subfig}
\usepackage{booktabs}

\def\bz{\mathbf{z}}
\def\h{\mathbf{h}}

\def\s{\mathbf{s}}
\def\bb{\mathbf{b}}

\def\u{\mathbf{u}}

\allowdisplaybreaks

\newcolumntype{?}{!{\vrule width 1.5pt}}
\usepackage{resizegather}
\usepackage{tikz}
\usepackage{tikz-qtree}
\newcommand*\circled[1]{\tikz[baseline=(char.base)]{
            \node[shape=circle,draw,inner sep=1pt] (char) {#1};}}
            
\newenvironment{itemize*}%
  {\begin{itemize}%
    \setlength{\itemsep}{0pt}%
    \setlength{\parskip}{0pt}}%
  {\end{itemize}}
  \newenvironment{enumerate*}%
  {\begin{enumerate}%
    \setlength{\itemsep}{0pt}%
    \setlength{\parskip}{0pt}}%
  {\end{enumerate}}

\usepackage{url}

\aclfinalcopy 

\title{Top-Down Tree Structured Text Generation}

\author{Qipeng Guo $^*$ \qquad Xipeng Qiu $^*$ \qquad Xiangyang Xue $^*$ \qquad Zheng Zhang $^\dagger$ \\
\hspace{-10mm} $^*$ School of Computer Science, Fudan University \\
\hspace{-10mm} $^\dagger$ New York University Shanghai \\
{\tt \normalsize \{qpguo16, \hspace{-2mm}xpqiu, \hspace{-2mm}xyxue\} @fudan.edu.cn \qquad zz@nyu.edu} \\}
\date{}

\begin{document}
\maketitle
\begin{abstract}
Text generation is a fundamental building block in natural language processing tasks. Existing sequential models performs autoregression directly over the text sequence and have difficulty generating long sentences of complex structures. This paper advocates a simple approach that treats sentence generation as a tree-generation task. By explicitly modelling syntactic structures in a constituent syntactic tree and performing top-down, breadth-first tree generation, our model fixes dependencies appropriately and performs implicit global planning. This is in contrast to transition-based depth-first generation process, which has difficulty dealing with incomplete texts when parsing and also does not incorporate future contexts in planning. Our preliminary results on two generation tasks and one parsing task demonstrate that this is an effective strategy.
\end{abstract}

\input{introduction.tex}

\input{background.tex}
\input{model.tex}

\input{parse.tex}
\input{train.tex}
\input{evaluation.tex}

\input{related-work.tex}
\input{conclusion.tex}

\bibliographystyle{acl_natbib_nourl}
\bibliography{main,ijcai18}

\end{document}

%% file: introduction.tex
\section{Introduction}
\label{sec-intro}


Generating coherent and semantically rich text sequences is crucial to a wide range of important natural language processing applications. Rapid progress is made using deep neural networks. In image captioning task, for example, a recurrent neural network (RNN) generates a sequence of words conditioned on features extracted from the image~\cite{vinyals2015show,xu2015show}. The same is true for machine translation, except the RNN is conditioned on the encoding of the source sentence~\cite{bahdanau2014neural}. The same framework can be extended to parsing to output a linearised parse tree~\cite{vinyals2015grammar}.

Despite these encouraging results, generating long sentences with complex structures remains an unsolved problem. The main issue is that the most popular models are sequential in nature, letting the RNN preform autoregression directly on text sequence. RNN has difficulty to remember and distinguish the complex and long dependencies embedded in a flattened sequence. This is true even with advanced variants such as LSTM and GRU~\cite{hochreiter1997long, chung2014empirical}. Error signals either can not back-propagate to the right sources or worse yet do so indiscriminately, leading to overfitting. Attentional mechanism~\cite{bahdanau2014neural} sidesteps this issue but is applicable only when alignment is possible (e.g. machine translation).

Sentences are inherently hierarchical. Properly leveraged, hierarchical structures in the form of constituency parse tree or dependency tree can guide and scale text generation. Transition-based models~\cite{titov2010latent, buys2015bayesian, dyer2016recurrent} is a step towards this direction. A parser works from bottom-up, consuming a sentence while building the hierarchy. This working mechanism can be converted to a generative model, by replacing the SHIFT operator with an action that generates a word. We observe two difficulties here. One is that the bottom-up parsing actions can have difficulty dealing with partially complete texts. Second, the depth-first nature ignores the opportunity to have a global planning with future contexts.

This paper advocates a simpler approach. Our model, called Top-Down Tree Decoder (TDTD), treats sentence generation as a tree-generation problem. TDTD generates a constituent syntactic tree layer by layer until reaching the leaf nodes, at which points words are predicted. By modeling structures explicitly with a set of internal RNNs that summarize prediction histories among siblings and ancestors, error signals modify the model appropriately. Importantly, the breadth-first nature pushes the model to consider (a summary) of future contexts during the generation process. This form of regularization implicitly leads to a more global planning, instead of the purely local decision as in a sequential model.
We also give an extended version TDTD-P as a parser using an encoder-decoder framework.

These models are easy to train, only involving curriculum learning~\cite{bengio2009curriculum} and scheduled sampling~\cite{bengio2015scheduled}. Experiment results on two language generation tasks and one parsing task demonstrate the promise of this approach.

%% file: background.tex
\section{Probabilistic Text Generation}
\label{sec-bg}
We review two related approaches of probability text generation. They share the goal of generating a text sequence $X_{1:L}=x_1,x_2,\cdots,x_L$ where $x_i$ belongs to some vocabulary $\mathcal{V}$. A parametric model is learned from a dataset $\{X^{(n)}\}_{n=1}^N$,
typically with maximum likelihood method.

For each sentence $X^i$ we can always find a \emph{constituency tree}, or a \emph{parse tree} $T^i$, whose leaf nodes are the words in the sentence $X^i$ itself.

\subsection{Sequential Text Generation}


Observing that the joint probability of the sequence $X_{(1:L)}$ can be decomposed by chain rule: 
\begin{align}
p_\theta(X_{1:L}) = \prod_{i=1}^{L} p_\theta(x_{i}|x_{0:i-1}),
\end{align}
We can have a sequential generator which outputs a word $x_{i}$ at time step $i$, conditioning on all the historical words $x_{0:i-1}$, where $x_0$ is a special \emph{beginning-of-sentence} symbol and $\theta$ is the model.  With a recurrent network, we can summarize the history $x_{0:i-1}$ into a hidden state $\h_i$:
%
\begin{align}
\h_i = f(\h_{i-1},x_{i-1}),
\end{align}
and predict the distribution of $x_i$ by projecting $\h_i$ through a softmax operator.

This model is conceptually simple and works very well for short sentences. It, however, ignores syntactic structures embedded in a sentence.
To fix this, the parse tree can be deterministically linearized (e.g via depth-first traversal) into a modified sequence with words and constituencies.
The challenge with this approach is dealing with long dependency spans, especially for long and complex sentences.
Attentional mechanisms sidestep the issue in machine translation task by aligning hidden states of the decoder to hidden states of the encoder~\cite{bahdanau2014neural,vaswani2017attention}. While effective, it is clearly task-specific, applicable only in an encoder-decoder framework and when source sentences are available.

\subsection{Transition-based Tree Structure Text Generation}

Recent works to generate text by incorporating syntactic information include ~\cite{titov2010latent,buys2015bayesian,dyer2016recurrent}. They leverage transition-based parsing, where a syntactic tree is translated to an oracle sequence of actions, involving SHIFT, REDUCE and NT (open nonterminal) operations. These operations perform on an input buffer and a stack. Replacing the input buffer with an output buffer and a SHIFT operation with GEN($x$), these models can be modified as a generator ~\cite{sagae2005classifier,nivre2008algorithms}.


These methods explicitly model syntactic and hierarchical relationships among words. 
One problem is their complexity, including both the data structures to be maintained and the operations to be performed. For the latter, there are two kinds: the bottom-up ones to build a partial tree based on parsing actions, and the top-down ones to generate the next word. Parsing actions are hard to predict when the generated text is partial and incomplete. Also, the depth-first order of actions limits the opportunity to perform global planning.


%% file: model.tex
\section{Proposed Model}\label{sec-model}

Our core idea is straightforward. Given that the syntactic structure is a tree and is available in training set, our generation model simply focuses on generating a tree. This is done in a breadth-first fashion, ensuring the access to a summary of future context when needed.
Recall that the leaf nodes of a tree $T$ is simply the sequence $X_{1:L}$ itself, therefore $p(X_{1:L},T) = p(T)$.

Let $V^d$ be the sequence of nodes with length $l_d$ at layer $d$, i.e.:
\begin{align}
    V^{d} = v_1^{d},v_2^{d},\cdots,v_{l_d}^{d}.
\end{align}
We can rewrite the joint probability $p(T)$ via chain rule, but across depth:
\begin{align}
p(T)=\prod_{d=0}^{D-1} p(V^{d+1}|V^{d},\cdots,V^{1}),
\end{align}
where $D$ is the maximum depth of tree $T$. 

We now proceed to explain how to generate the tree layer-by-layer; \textrm{RNN} and \textrm{SM} denote the uni-directional RNN and softmax operator, respectively, each with their own learnable parameters.

\begin{figure}
    \centering
    \includegraphics[width=\linewidth]{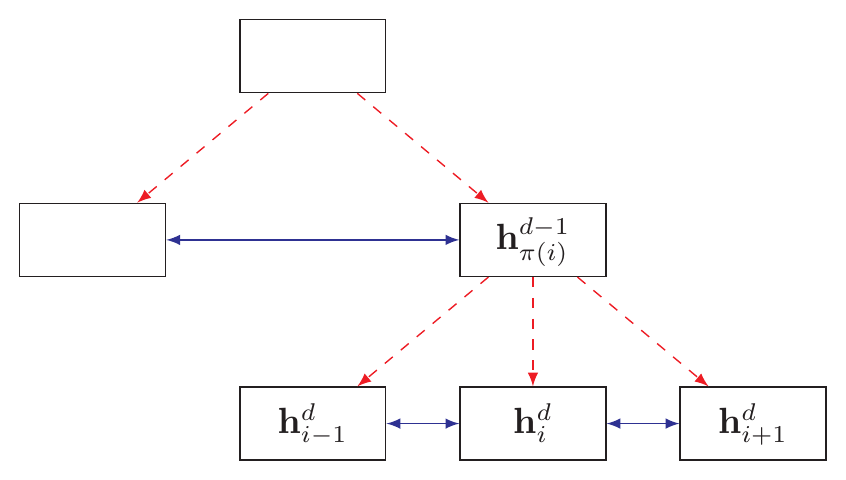}
    \caption{Tree-Stacked RNN.}
    \label{fig:encoder}
\end{figure}

To generate the nodes at $(d+1)$-th layer, we first encode the history information from the 1st layer to the $d$-th layer, $V^{1},\cdots,V^{d}$ with a tree-stacked RNN to encode the history information (see Figure~\ref{fig:encoder}). Here, each layer is encoded by a  bidirectional RNN to compute the context representation. For each node $v^{d}_i$,
let $\pi(i)$ denote the index of the parent node and $v_{\pi(i)}^{d-1}$ be its parent node.
The hidden state $\h^{d}_i$ of node $v^{d}_i$ is then:
\begin{align}
\overrightarrow{\h}_i^{d} &= \textrm{RNN}(\overrightarrow{\h}_{i-1}^{d},\mathbf{v}^{d}_i,\h_{\pi(i)}^{d-1}),\\
\overleftarrow{\h}_i^{d} &= \textrm{RNN}(\overleftarrow{\h}_{i+1}^{d},\mathbf{v}^{d}_i,\h_{\pi(i)}^{d-1}),\\
\h_i^{d}&=[\overrightarrow{\h}_i^{d};\overleftarrow{\h}_i^{d}],
\end{align}%
where $\mathbf{v}^{d}_i$ is the embeddings of node $v^{d}_i$, $\h_{\pi(i)}^{d-1}$ is the hidden state of the parent of node $v^{d}_i$.

Unlike a vanilla stacked RNN, the input of RNN at node $v^{d}_i$ in tree-stacked RNN is the output of the RNN at node $v_{\pi(i)}^{d-1}$, the parent of $v^{d}_i$.

Thus, the hidden state $\h_i^{d}$ is a contextual representation of node $v_i^{d}$ by integrating its sibling and ancestral information.
Therefore, $\mathbf{H}^{d}=[\h_1^{d},\h_2^{d},\cdots,
\h_{l_d}^{d}]$ of all the nodes in $(d)$-th layer captures all the information of $V^{d},\cdots,V^{1}$.
The conditional probability of the nodes in $d\mathrm{+}1$-th layer is

{\small
\begin{align}
    p(V^{d+1}|\mathbf{H}^{d}) & = \prod_{i=1}^{l_{d+1}} p(v_i^{d+1}|v_{i-1}^{d+1},\cdots,v_{1}^{d+1},\mathbf{H}^{d}).
\end{align}}%

Let $\u^{d+1}_i$ encode all the information of $v_{i-1}^{d+1},\cdots,v_{1}^{d+1}$, the left siblings or cousins of $v_i^{d+1}$. We have:
\begin{align}
    p(V^{d+1}|\mathbf{H}^{d}) &=\prod_{i=1}^{l_{d+1}} p(v_i^{d+1}|\u_{i}^{d+1},\mathbf{H}^{d})\\
    &=\prod_{i=1}^{l_{d+1}} p(v_i^{d+1}|\u_{i}^{d+1},\h_{\pi(i)}^{d}),
\end{align}
where $\h^{d}_{\pi(i)}$ encodes the parent of $v_i^{d+1}$.

Generating node $v^{d+1}_i$ depends on the prediction history on its siblings and ancestors in the tree, as we describe next.


\paragraph{Gen-RNN} We first integrate the prediction history of the siblings or cousins into $\u_{i}$ by a forward RNN,
\begin{align}
    \u^{d+1}_i &= \textrm{RNN}(\u^{d+1}_{i-1},v_{i-1}^{d+1}),
\end{align}

Then $\u^{d+1}_i$ is concatenated with history of the ancestor $\s_{\pi(i)}^{d-1}$ to form a representation for prediction.
{\small
    \begin{align}    
    p(y_\mathcal{V}|\u_{i}^{d+1},\s_{\pi(i)}^{d})
    &=\textrm{SM}(W_{\mathcal{V}} [\u_{i}^{d+1}, \s_{\pi(i)}^{d}]+\bb_{\mathcal{V}}), \\
    p(y_\mathcal{N}|\u_{i}^{d+1},\s_{\pi(i)}^{d})
    &=\textrm{SM}(W_\mathcal{N} [\u_{i}^{d+1}, \s_{\pi(i)}^{d}]+\bb_\mathcal{N}).
    \end{align}
}%

Terminal (with subscript $\mathcal{V}$) and nonterminal (with subscript $\mathcal{N}$) words are predicted separately because the later is much smaller set. Finally, we also predict a special STOP symbol, identifying the point when the model switches to a cousin or next layer after the last symbol of the current layer. The process continues layer by layer until all leaf nodes are terminal symbols.

\subsection{Analysis}
Our proposed model has the following advantages:
\begin{itemize*}
    \item Compared to the sequential models, the depending paths are usually significantly shorter in the tree structure. By explicitly encoding the path, the model also has less unrelated connections, reducing overfitting. 
    
    \item The top-down process enforces future planning as a form of regularization. Node splitting in higher level pushes the model to plan the production of the whole sentence rather than focus on local features, as typically is the case in sequential models.
    
    \item Compared to the transition-based models, our model does not need to incorporate bottom-up parsing operations, which are usually difficult for a partially generated text. Like sequential models, a transition-based model also lacks the overall planning due to its depth-first nature.
    
    \item We employ constituent syntactic tree to generate text in this paper; adapting dependency tree is not difficult. 
\end{itemize*}

%% file: parse.tex
\section{TDTD-P: an Extension to Generative Parsing Model}\label{sec-parse}

\begin{figure*}
    \centering
    \includegraphics[width=0.9\textwidth]{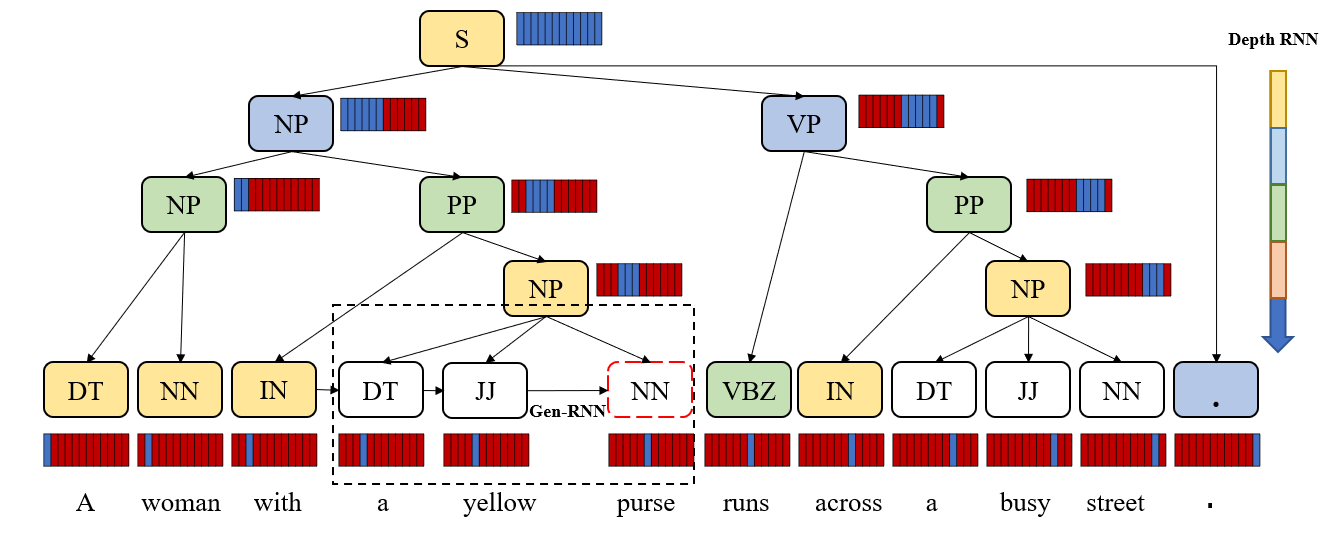}
    \caption{An overall view of our parsing model, TDTD-P. The input of the root node is a vector output by a Bi-directional RNN encoder which takes the texts (the sentences in the bottom of this figure) as inputs. The red-blue bar represents the attention over the input sequence. The content in the black dash box shows the pipeline of generating the red dash node (NN), the model merges three kinds of information to get the vector representation and its label. We use Gen-RNN to collect the contextual information from its left siblings and ancestors, and the attention results over the input sequence. The whole tree is growing in a Top-Down Breadth-First fashion, such as \{S,NP,VP,NP,PP,VBP,PP,.. \}. }\label{fig-model-parsing}
\end{figure*}

A parsing model receives a flat word sequence $X_{1:T}=x_1,x_2,\cdots,x_L$ as input and outputs its constituency parse tree $T$ to represent the compositional relationships among these words.

To extend our proposed model to a generative parsing model, we formulate parsing as a conditional tree generation, where the leaf nodes must exactly match the input $X_{1:L}$.

{\small\begin{align}
p(T|X_{1:L}) = \prod_{d=0}^{D-1} p(V^{d+1}|V^{d},\cdots,V^{1}|X_{1:L}),
\end{align}}%

Therefore, the generative parsing can be regarded as an encoder-decoder problem. We use bidirectional RNN to encode the input sentence, and then apply our model as a decoder to generate the parse tree.

\paragraph{Attention Mechanism}
Since the input sequence can be very long ($>40$ tokens), the bidirectional RNN encoder can miss details. A popular solution is using attention mechanism to access the origin sequence and perform alignment dynamically at each time-step. We also apply this idea here.

Assuming that the encoding of input sentence is $\mathbf{H}^{enc}=\h_{1}^{enc}, \h_{2}^{enc}..., \h_{L}^{enc}$, to generate the $i$-th node of $d$-th layer, we use the current state $[\u_{i}^{(d)}; \h_{\pi(i)}^{(d-1)}]$ to obtain an aligned context vector $\h^{(att)}$ from the input sequence:
\begin{align}
\h^{(att)} = \textrm{attention}([\u_{i}^{d+1}; \h_{\pi(i)}^{d}],\mathbf{H}^{enc}),
\end{align}
where $\textrm{attention}(\cdot)$ is an dot-production attention function \cite{vaswani2017attention}.

Then the context vector $\h^{(att)}$ is used as an additional feature to generate the next node,
\begin{align}
\bz &= [\u_{i}^{d+1}; \h_{\pi(i)}^{d};\h^{(att)}],\\
p(v_i^{d+1}|\u_{i}^{d+1},\h_{\pi(i)}^{d})
&=\textrm{SM}(W \bz +\bb).
\end{align}


Since the generation process of TDTD is top-down breadth-first, it is hard to make the leaf nodes exactly match the input $X_{1:L}$. In this paper, we treat our TDTD-P model as a scorer for measuring the matching of a candidate tree and its text. We use scores from TDTD-P to re-rank candidate trees and select the tree which has the highest score as the parsing result (see Sec-\ref{sec-exp-parse}).

%% file: train.tex
\section{Training} \label{sec-train}

Given a training set $\{T^{(n)}\}_{n=1}^N$, we use mini-batch stochastic gradient descent to look for a set of parameters $\theta$ that maximizes the log likelihood of producing the correct tree.
\begin{align}
\theta &= \arg\max \frac{1}{N}\sum_{n=1}^{N} \log p_\theta(T^{(n)}),
\end{align}
where $\theta$ denotes all the parameters in our model.


We employ two techniques to improve training for the parsing task. The first is curriculum learning~\cite{bengio2009curriculum}: we constrain the depth and width of the tree to limit the solution space at the beginning of training and gradually relax it. The second is schedule sampling which has been shown to be beneficial dealing with distribution shift in sequence learning~\cite{bengio2015scheduled}: the labels are gradually replaced by model predictions with a slowly annealing probability, following a greedy strategy.



%% file: evaluation.tex
\section{Experiments}\label{sec-exp}

To evaluate our model, we conduct experiments on two generation tasks. The first generates trees, using a synthetic tree dataset produced by an oracle of probabilistic context-free grammar (PCFG) model. The second generates text sentences, learning from a large movie review corpus. Additionally, we test the parsing task on PTB.

\paragraph{System compared} We compare against a similarly configured vanilla LSTM, SeqGAN and LeakGAN~\cite{yu2017seqgan,guo2017long}; all of them are sequential generator based on the recurrent neural network. For these models, we first linearise the tree structure to brackets expression form. SeqGAN and LeakGAN sidestep exposure bias by adopting an adversarial training paradigm, and LeakGAN optimizes further for long sequences. The GAN approach, however, suffers from mode collapsing, as our results show.

\paragraph{Comparison methodology} It is difficult to measure the quality of a generated text with syntactic structure. First, if we evaluate our model as joint generative model $p(X,T)$, there is no oracle to judge the quality of tree $T$. Second, if we evaluate our model as a language model, the marginal probability $p(X) = \sum_{T} p(X,T)$ is intractable. Therefore, we conduct two experiments. The first experiment uses an oracle to score the generated tree, and the second uses BLEU score to judge the generated text regardless of its syntactic tree.

\paragraph{Implementation Details}

In TDTD, both the Depth-RNN and the Gen-RNN are single-layer GRU, and the Layer-RNN is a single-layer bidirectional GRU. To keep the setting comparable to previous works, we set the hidden size to 32 and the size of input embedding of tags to 32. In the TDTD-P variant, we set the hidden and embedding size to 128.





\subsection{Exp-I: Synthetic Data}

\begin{table}\centering
    \begin{tabular}{cccc|c}
    \toprule
    X & $\rightarrow$ & \multicolumn{2}{c|}{Y} & prob \\
    \hline
    $\text{S}_{18}$ & $\rightarrow$ &  \multicolumn{2}{c|}{$\text{VP}_{30}$} & 0.977 \\
     $\text{ADJP}_{21}$ &  $\rightarrow$ & \multicolumn{2}{c|}{$\text{JJ}_{37}$} & 0.959 \\
     $\text{@CONJP}_{0}$ & $\rightarrow$ & $\text{RB}_{24}$ & $\text{RB}_{11}$ & 0.954 \\
     $\text{NP}_{13}$ & $\rightarrow$ & $\text{DT}_1$ & $\text{NN}_{42}$ & 0.907 \\
     $\text{PP}_7$ & $\rightarrow$ & $\text{IN}_{28}$ & $\text{NP}_{21}$ & 0.845 \\
     $\text{PP}_{14}$ & $\rightarrow$ & $\text{TO}_0$ & $\text{NP}_{42}$ & 0.629 \\
     $\text{S}_{16}$ & $\rightarrow$ & $\text{NP}_{42}$ & $\text{VP}_{30}$ & 0.540 \\
     $\text{ADVP}_{13}$ & $\rightarrow$ & $\text{NP}_{13}$ & $\text{RB}_{43}$ & 0.386 \\
     $\text{NP}_{17}$ & $\rightarrow$ & $\text{CD}_7$ & $\text{NN}_{16}$ & 0.351 \\
     \hline
    \end{tabular}
    \caption{PCFG production rule examples from Berkeley parser.}
    \label{tab-pcfg-examples}
\end{table}


We conduct a simulated experiment with synthetic data, using an oracle PCFG model to play two roles: generating training samples and evaluating the syntactic trees generated by our model. Berkeley Parser~\cite{petrov2006learning} has both capabilities. Its PCFG model contains around $1.9M$ production rules and can generate various constituency trees, available on-line\footnote{https://github.com/slavpetrov/berkeleyparser}. These rules and their associated probability can also evaluate the likelihood of newly generated samples (see Table~\ref{tab-pcfg-examples}).


To ensure stability, we remove production rules with a probability less than $1\textrm{e}^{-6}$. Likewise, in the evaluation, a $1\textrm{e}^{-6}$ penalty is imposed for unseen rules in our samples. Since we wish the generated text to be a complete sentence instead of a phrase or substructures, we generate samples by limiting the starting nodes to be a subset consisting of the non-terminals ``S$_*$'' family. In addition, The maximum depth of generated trees is set to $7$.




Under the above configurations, we prepare three datasets, varying number of nodes as 10, 15 and 20. For each dataset, we sampled 10,000 trees as the training set. This way, we can inspect how our model scales with sentence length. Note that linearisation increases sequence length: samples in brackets form are always three times of the number of nodes.




\paragraph{Evaluation}

We use the negative log-likelihood (NLL) to evaluate the generated samples by the oracle PCFG model. As mention before, we limit the starting nodes to be the ``S$_*$'' subset. Therefore, for a tree $T$ with root $v_1$, we can use the conditional probability $P(T|v_{1})$ instead of $P(T)$.

Since the oracle is a PCFG model, the probability can be decomposed by a chain of productions. Thus, NLL of a given tree $T$ is:
\begin{align}
 NLL(V| v_{1}) & = -\log P_{oracle}(T | v_{1}) \\
 &= - \sum_{v_i \in T} \log P_{oracle}(V_i^C | v_i),
\end{align}
where $v_{1}$ is the root node, $V_i^C$ are the children of the non-leaf node $v_i$, and $P_{oracle}(V_i^C | v_i)$ is the probability of production $v_i\rightarrow V_i^C$ in oracle.


\begin{table}[t!]\centering
\setlength{\tabcolsep}{5pt}
    \begin{tabular}{lcccc}
    \toprule
    Model & Len & NLL & Fail (\%) & Dup (\%) \\
    \midrule
    Oracle & $10(30)$ & $2.43$ & $-$ & $11.2$ \\
    LSTM & $10(30)$ & $3.85$ & $54.1$ & $8.6$ \\
    SeqGAN & $10(30)$ & $0.67$ & $2.6$ & $93.0$ \\
    LeakGAN &$10(30)$ &$8.25^{\dagger}$ &$52.0$ &$0.0$ \\
    TDTD & $10(30)$ & $\mathbf{3.58}$ & $0.0$ & $25.7$ \\
    \midrule
    Oracle & $15(45)$ & $2.63$ & $-$ & $1.3$ \\
    LSTM & $15(45)$ & $6.39$ & $66.2$ & $0.0$ \\
    SeqGAN & $15(45)$ & $7.41^{\dagger}$ & $93.7$ & $0.0$ \\
    LeakGAN &$15(45)$ &$6.42^{\dagger}$&$78.2$ & $0.0$ \\
    TDTD & $15(45)$ & $\mathbf{3.86}$ & $0.0$ & $16.7$ \\
    \midrule
    Oracle & $20(60)$ & $2.85$ & $-$ & $0.3$ \\
    LSTM & $20(60)$ & $7.55$ & $67.8$ & $0.0$ \\
    SeqGAN & $20(60)$ & $7.88^{\dagger}$ & $94.2$ & $0.0$ \\
    LeakGAN &$20(60)$ & $6.79^{\dagger}$ & $71.1$ & $0.0$ \\
    TDTD & $20(60)$ & $\mathbf{4.32}$ & $0.0$ & $11.8$ \\
    \bottomrule
    \end{tabular}
    \caption{Results on synthetic datasets. ``Len'' means number of nodes in the tree, and the number in bracket is the length of brackets sequence. ``Fail'' denotes the percentage of ill-formed generated samples, which have unmatched brackets and cannot be converted into a tree. ``Dup'' is the percentage of duplicate samples in all the generated samples. Failed samples are not counted in the NLL score. $\dagger$ means that the performance falls after a few iterations during the training, in which case we perform early-stop.}
    \label{tab-gen}
\end{table}

\paragraph{Generation results}

Table \ref{tab-gen} shows that, except SeqGAN with 10 nodes, our model outperforms all others on NLL. However, SeqGAN suffers from mode collapsing, as seen by its high duplication rate in its generated samples (93\%, the $3$-rd row). The performance of all sequential models drop uniformly with longer sentences and become increasingly difficult to recover coherent tree structures. These results indicate the fundamental defect of the sequential models.

TDTD works consistently well on small and large trees, suggesting that the hierarchical structure has a high potential for dealing with long sequence because the dependency path of tree structures grows much slower.





\subsection{Exp-II: Real-world Text Generation}

To evaluate the ability to generate real-world texts, we experiment our method on an unconditional text generation task similar to ~\cite{yu2017seqgan,guo2017long} but use a large IMDB text corpus ~\cite{diao2014jointly} to train our model. This dataset is a collection of 350K movie reviews and contains various kinds of compound sentences.
We select sentences with the length between 17 and 25, set threshold at 180 for high-frequency words and only select sentences with words above that threshold.
Finally, we randomly choose 80000 sentences for training and 3000 for testing, with vocabulary size at 4979 and the average sentence length at 19.6. We use Stanford Parser to obtain the syntactic tree for training our model.

We use Stanford Parser\footnote{https://nlp.stanford.edu/software/} to obtain the syntactic tree, and BLEU score \cite{papineni2002bleu} to measure similarity degree between the generated texts and the texts in test set.

\begin{table}[t!]\centering
\setlength{\tabcolsep}{2pt}
    \begin{tabular}{lcccc}
    \toprule
    & LSTM & SeqGAN & LeakGAN & TDTD \\
    \midrule
    BLEU-2 & 0.652 & 0.683 & \bf 0.809 & 0.718\\
    BLEU-3 & 0.405 & 0.418 & 0.554 & \bf 0.568\\
    BLEU-4 & 0.304 & 0.315 & 0.358 & \bf 0.375 \\
    BLEU-5 & 0.202 & 0.221 & 0.252 & \bf 0.263\\
    \bottomrule
    \end{tabular}
    \caption{BLEU results on IMDB.}\label{tb:imdb}
\end{table}

The results in Table \ref{tb:imdb} show that the BLEU scores of TDTD  are consistently higher than the other models except BLUE-2. The results indicate that the generated sentences of TDTD are of high quality to mimic the real text.

\begin{table}[t!]\centering
    \small\setlength{\tabcolsep}{2pt}
    \begin{tabularx}{1\linewidth}{l|X}
    \hline Method & Cases \\
    \hline \multirow{5}{*}{LSTM} & (1) His monster is modeled after the old classic ``B'' science fiction movies we hate to love, but it was known with the first humor .\\
    & (2) But I was completely bored the movie the way I saw this movie at the same time . \\
    \hline \multirow{6}{*}{SeqGAN} & (1) This does not star Kurt Russell, but rather allows him what amounts to an extended cameo . \\
    & (2) Don't all of them because it will not be a very boring love stories and so many people judge at some people . \\
    \hline \multirow{6}{*}{LeakGAN} & (1) The film is a simple, the premise that an ordinary guy can become a hero, only if he can be easy to see . \\
    & (2) It is a non-stop adrenaline rush from beginning to end, and as for how it was the main villain . \\
    \hline \multirow{4}{*}{TDTD} & (1) Need for Speed is a great movie with a very enjoyable storyline and a very talented cast . \\
    & (2) The story is modeled after the old classic ``B'' science fiction movies \\
    \hline
    \end{tabularx}
    \caption{Samples from different models}
    \label{tab:generated_samples}
\end{table}

\paragraph{Case Study}

Table~\ref{tab:generated_samples} shows some generated samples of our and other methods, which also shows that the generated sentences of TDTD are of higher global consistency and better readability than others.

\begin{figure}[t!]\centering
\begin{tikzpicture}[scale=0.6]
\tikzset{frontier/.style={distance from root=200pt}}

\Tree [.S [.NP [.DT The ] [.NNS effects ] ] [.\textcolor[rgb]{1.00,0.00,0.00}{VP} [.VP [.VP [.VBP are ] [.\textcolor[rgb]{1.00,0.00,0.00}{NP} [.\textcolor[rgb]{1.00,0.00,0.00}{NP} [.NN nothing ]]]] [., , ] [.CC but ] [.VP [.VBP are ] [.ADVP [.RB still ] ] ] ] [., , ] [.S [.NP [.\textcolor[rgb]{1.00,0.00,0.00}{DT} \textcolor[rgb]{1.00,0.00,0.00}{all} ] ] [.VP [.VBG considering ] ] ] ] [.. . ] ]
\end{tikzpicture}
\caption{An error case generated by TDTD, involving three main errors.
}
\label{fig:tree_gen_example}
\end{figure}
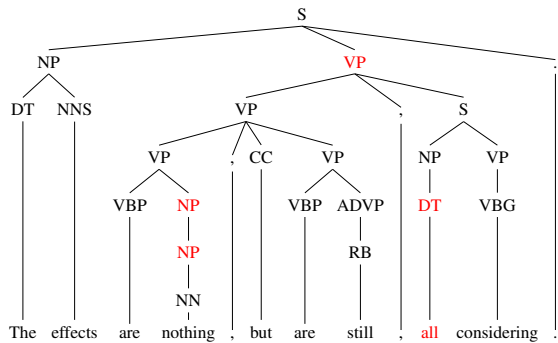

We also give an error analysis in Figure~\ref{fig:tree_gen_example}, which illustrates a real generated text with its syntactic tree, including some errors. We make a few observations:
\begin{itemize}
    \item Our model successfully applies global planning early on, as expected. Such as ``S $\rightarrow$  NP VP .'' at the top-level. 
    \item This example highlights some of the error patterns, some of them were contributed by bad trees that were automatically parsed on the noisy IMDB texts. In this example, the first error (VP $\rightarrow$ VP, ``,'' , S) occurs in the earlier phase, which persists till the end. 
    The second error is the redundant double ``NP''s, and the third error is that the role of ``all'' is closer to pronoun rather than a determiner in this sentence.
\end{itemize}



\subsection{Exp-III: Parsing results}
\label{sec-exp-parse}
Although our model is not intended for parsing, it can be converted into a generative parser by adopting an encoder-decoder structure (TDTD-P; see Section~\ref{sec-parse}). The experiment is conducted on Penn Treebank, $\S$2-21 are used for training, $\S$24 used for development, and $\S$23 used for evaluation.

Since this parser is a generative model, we adopt the same evaluation paradigm of RNNG ~\citep{dyer2016recurrent}. We firstly obtained 100 independent candidates\footnote{The samples are publicly available on https://github.com/clab/rnng, released by \cite{dyer2016recurrent}.} for each test case with a discriminative parser.
Then we re-rank the candidates according to their probabilities computed by our model.


We use the F1-score, a standard measurement in parsing tree evaluations as our metric. Table~\ref{tb:parsing} gives the performance of our model, along with several representative models. RNNG-D (Discriminative) is the result before re-ranking, and RNNG-G is the re-ranked results by RNNG. GA-RNNG incorporates gated attention into RNNG. The result shows that TDTD-P can work as a decent parser, but does not work as well as RNNG methods. One reason is that TDTD-P only has top-down actions, unlike RNNG that also include bottom-up parsing actions. Therefore TDTD-P is less suited for parsing task.

\begin{table}[t!]\centering\setlength{\tabcolsep}{15pt}
    \begin{tabular}{lc}
    \toprule
    Algorithm & F1 \\
    \midrule
\citet{petrov2006learning}    & 90.4\\
\citet{zhu2013fast}  &  90.4\\
\citet{zhu2013fast} $^{\dagger}$ & 91.8\\
\citet{vinyals2015grammar} – PTB only &88.3\\
\citet{vinyals2015grammar} $^{\dagger}$ & 92.1\\
    \midrule
    RNNG-D \cite{dyer2016recurrent} & 91.2 \\
    RNNG-G \cite{dyer2016recurrent} & 93.3 \\
    GA-RNNG \cite{kuncoro2016recurrent}  & 93.5 \\
    TDTD-P & 91.9 \\
    \bottomrule
    \end{tabular}
    \caption{Parsing results on PTB. $^{\dagger}$ means semi-supervised method.}\label{tb:parsing}
\end{table}

\begin{figure*}[t]
\centering
  \subfloat[A parsing result.]{\label{fig:parse_example}
\begin{tikzpicture}
\tikzset{frontier/.style={distance from root=150pt}}
\Tree [.S [.CC But ] [.NP\circled{1} [.PRP he ] ] [.VP\circled{1} [.MD could ] [.RB n't ] [.VP\circled{2} [.VB sell ] [.NP\circled{2} [.\textcolor{red}{DT} any ] ] ] ] [.. . ]
]
\end{tikzpicture}\hspace{3em}
}
\subfloat[Visualization of attention.]{
\includegraphics[width=0.4\textwidth]{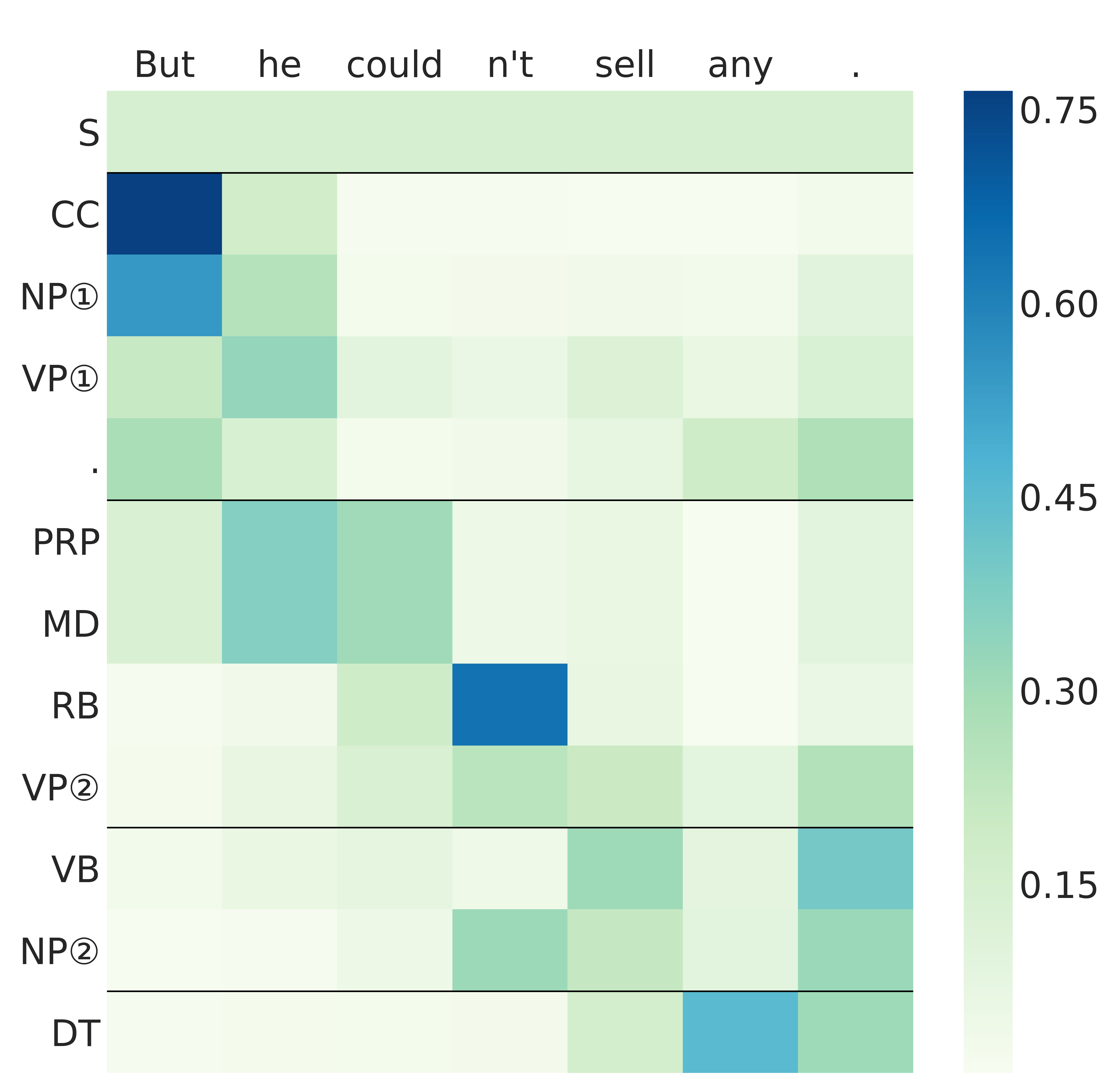}
}
\caption{A case study of TDTD-P. (a) The red token indicates an error in the prediction, our model predicted a wrong label ``NN'' and the correct label is ``DT''. Circled numbers are used to distinguish tags with the same name. (b) Visualization of attention, and horizontal lines split the nodes with its depth.}\label{fig:pars-vis}
\end{figure*}

\paragraph{Case Study}
In the parsing experiment, we are interested in how our model converts a flatten sequence into a tree, a meaningful viewpoint is probing the attention of each node. Specifically, we look for the attention difference between sibling nodes and the difference between the parent and the children nodes. Figure~\ref{fig:pars-vis} gives an illustration of how attention working in TDTD-P.

%% file: related-work.tex
\section{Related Work}
\label{sec-related}

Although deep neural networks have made a great progress in text generation, most of them employed sequential models, performing autoregression directly on the text sequence. Generating text by incorporating its syntactic tree structure has not been a popular approach.

\citet{vinyals2015grammar} linearises parsing trees to brackets expression form and use the attention-enhanced sequence-to-sequence model to parse sentences. Although their method can generate a tree-structured text with slight modification, the linearization process only aggravates the issue of long-distance dependency for sequential models. Also, the attentional mechanism is available in an encoder-decoder framework. 

\citet{dyer2016recurrent} proposes recurrent neural network grammars, a generative probabilistic model of phrase-structure trees. Their model operates via a recursive syntactic process reminiscent of probabilistic context-free grammar generation. However, decisions parametrized with RNNs are conditioned on the entire syntactic derivation history. This greatly relaxes context-free independence assumptions.

All the above methods generate text in depth-first traversal order. Different from them, ours is the first work to use syntactically structured neural models to generate language in top-down breadth-first fashion.

Besides, there are also some works that explore the syntactically structured neural architectures in a number of other applications, including discriminative parsing~\cite{socher2011parsing}, sentiment analysis~\cite{socher2013recursive,tai2015improved}. However, these model focus to learn sentence representation and they typically utilize the syntactical structure in a bottom-up fashion. 

%% file: conclusion.tex
\section{Conclusion}
\label{sec-conclusion}
In this paper, we propose a new method that treats text generation as a tree-generation problem. The approach explicitly utilizes syntactic information and considers a more global planning, two issues existing models fail to deal with efficiently. The results are promising, and future extensions include incorporating work from reinforcement learning and graph generation.  